\pdfoutput=1
\documentclass[10pt, a4paper]{article}
\usepackage{lrec2022} 
\usepackage{multibib}
\newcites{languageresource}{Language Resources}
\usepackage{graphicx}
\usepackage{tabularx}
\usepackage{caption}
\usepackage{soul}
\usepackage{titlesec}
\titleformat{\section}{\normalfont\large\bfseries\center}{\thesection.}{1em}{}
\titleformat{\subsection}{\normalfont\SmallTitleFont\bfseries\raggedright}{\thesubsection.}{1em}{}
\titleformat{\subsubsection}{\normalfont\normalsize\bfseries\raggedright}{\thesubsubsection.}{1em}{}
\renewcommand\thesection{\arabic{section}}
\renewcommand\thesubsection{\thesection.\arabic{subsection}}
\renewcommand\thesubsubsection{\thesubsection.\arabic{subsubsection}}

\usepackage{epstopdf}
\usepackage[utf8]{inputenc}

\usepackage{hyperref}
\usepackage{xstring}

\usepackage{color}

\newcommand{\secref}[1]{\StrSubstitute{\getrefnumber{#1}}{.}{ }}

\title{Quality and Efficiency of Manual Annotation: Pre-annotation Bias}

\name{Marie Mikulová, Milan Straka, Jan Štěpánek, Barbora Štěpánková, Jan Hajič} 

\address{Institute of Formal and Applied Linguistics \\
         Computer Science School, Faculty of Mathematics and Physics, Charles University, Prague \\
                  \{mikulova,straka,stepanek,stepankova,hajic\}@ufal.mff.cuni.cz\\}

\abstract{
This paper presents an analysis of annotation using an automatic pre-annotation for a mid-level annotation complexity task - dependency syntax annotation. It compares the annotation efforts made by annotators using a pre-annotated version (with a high-accuracy parser) and those made by fully manual annotation. The aim of the experiment is to judge the final annotation quality when pre-annotation is used. In addition, it evaluates the effect of automatic linguistically-based (rule-formulated) checks and another annotation on the same data available to the annotators, and their influence on annotation quality and efficiency. The experiment confirmed that the pre-annotation is an efficient tool for faster manual syntactic annotation which increases the consistency of the resulting annotation without reducing its quality.
\\ \newline  }
\usepackage[absolute]{textpos}
\begin{document}
\begin{textblock}{16}(0,0.1)\centerline{\small This paper was published in \textbf{LREC 2022} -- please cite the published version {\scriptsize\url{https://aclanthology.org/2022.lrec-1.312/}} instead.}\end{textblock}

\maketitleabstract

\section{Introduction}
\label{sec:intro}

Developing high-quality annotated corpora representing natural language phenomena that can be used by machine learning tools or explored by linguists solving various research tasks is time-consuming and expensive. This paper presents results of a carefully designed experiment that shed some light on the issue of efficiency vs. quality of manual syntactic annotation. For this purpose, we used manually annotated parts of the Prague Dependency Treebank - Consolidated, PDT-C in sequel  \citelanguageresource{lrPDT-C}\footnote{\url{https://hdl.handle.net/11234/1-3185}} under various experimental set-ups: annotation with no support (``from-scratch mode''), annotation supported by the use of automatic pre-annotation (using a high-accuracy parser), availability of other annotation on the same data, and ``online'' checking rules (implemented in such a way that the annotator can check the manual annotation or manual post-editing immediately). The experiments focus on dependency syntax annotation: each token of a sentence is assigned its head and specific type of the dependency relation.

With the experiment, we aim to answer three main research questions, the first of which is crucial:

\begin{itemize}
\item \textbf{Does the quality of manual annotation remain acceptably high if pre-annotation is used?} How much the manual correction of automatic pre-annotation increases quality? The concern is that annotators might tend to simply go along with the pre-annotation, which would lead to a lower quality than they could produce by a fully manual annotation (cf. similar question in  \newcite{rehbein-etal-2009-assessing}). 

\item What effect do other available tools (checking rules, other annotation on the same data) have on the quality and efficiency of the annotation and to what extent?

\item Which set-up is most useful for manual dependency syntax annotation? We want to find out how to best prepare the annotation environment and workflow for the syntactic annotation of 
2,000,000 tokens
in the PDT-C project (see Sect.~\ref{sec:task}) and possibly generalize the findings for other similar projects.
\end{itemize}

To evaluate the results, we use the usual methods for measuring accuracy and inter-annotator agreement in the dependency syntax tasks (UAS, LAS, Cohen's kappa). 

The experiment confirmed the usefulness of automatic pre-annotation for efficiency of annotation and quality of the result. The influence of the other support tools on efficiency and quality of annotation was not as positive as expected. However, when used together they significantly increase the quality of the annotation.

In the following Sect.~\ref{sec:related-work}, we summarize previous related work on similar tasks. In Sect.~\ref{sec:task}, the annotation task is briefly described. Annotation support tools used within the experiment (parser, automatic checks, availability of other annotation) are presented in Sect.~\ref{sec:tools}. The core of the article is Sect.~\ref{sec:experiment}, in which the experiment is described and the results are presented and evaluated. Conclusions and future work are summarized in Sect.~\ref{sec:conclusions}.

\section{Related Work}
\label{sec:related-work}

Various time-saving methods while maintaining data quality have already been explored. An overview of studies on this topic is given by \newcite{grouin2014optimizing}.  A number of previous works have demonstrated the usefulness of automatic pre-annotation by verifying it with various annotation experiments. 

The experiments were performed on different annotation tasks (POS annotation \cite{fort-sagot-2010-influence}, semantic role labeling  \cite{chou-etal-2006-semi,rehbein-etal-2009-assessing}, name entity annotation  \cite{grouin2014optimizing,rosset-etal-2013-automatic}, treebank construction \cite{marcus1993building,chiou-etal-2001-facilitating,gupta-etal-2010-partial,voutilainen2011double}) for different purposes (comparison of annotation quality by experienced and inexperienced annotators \cite{gupta-etal-2010-partial}, the effect of different types of pre-annotation, the difference between using a high-quality and lower-quality parser \cite{rehbein-etal-2009-assessing,chiou-etal-2001-facilitating}), which corresponds to the different set-ups of the experiments. Experimental design was chosen not only in line with research question that the authors wanted to address, but also in line with the number of annotators, the amount of data and time available.  

The previous works also confirmed that a clear and controlled, well performed experimental design is necessary for minimizing side effects on the results of the experiment (annotator influence, learning effect, inappropriate or/and different data; cf. the design of the experiment in \newcite{rehbein-etal-2009-assessing}, where an ongoing learning effect (repeated annotation of the same data by the same annotators) seems to have led to a distortion of the results).

The approaches that are closest to ours are that of treebank construction.  One of the earliest studies that investigated the usefulness of automatic pre-annotation for treebank construction was carried out by \newcite{marcus1993building}. In the context of the development of the Penn Treebank, they compared the semi-automatic approach to a fully manual annotation and found that the semi-automatic method resulted both in a significant reduction of annotation time, and in increased inter-annotator agreement and accuracy. \newcite{chiou-etal-2001-facilitating} investigated the effect of pre-annotation in the context of the development of the Chinese Penn Treebank. They experimented with two different parsers (using a high-accuracy and lower-accuracy parser) and found that both significantly shortened the total annotation time while maintaining accuracy.  \newcite{gupta-etal-2010-partial} described a semi-automatic approach to expedite the process of manual annotation of a Hindi Dependency Treebank. In addition to the influence of automatic pre-annotation, they also experimented with two groups of annotators: experienced and inexperienced. From their observations, it seems that a semi-automatic process is an effective way of doing dependency annotation when the annotators are trained and experienced. The experiment also confirmed the fact that treebank annotation is not a trivial task and supervision is required for carrying out the task in an efficient manner. \newcite{voutilainen2011double} documented a double-blind experiment on syntactic annotation of Finnish texts. 
The results suggest that as a result of reviews and negotiations, inter-annotator agreement can be much higher than the corresponding labeled attachment scores reported for state-of-the-art dependency parsers.

Metrics for measuring annotation quality have been discussed in \newcite{artstein2008inter}. In a more recent work, \newcite{skjaerholt-2014-chance} pointed out the inadequacy of existing metrics for syntactic annotation and proposes his own metric adapted to the structure of syntactic annotation. However, the UAS and LAS (unlabeled and labeled attachment scores, respectively) for measuring accuracy \cite{yang-li-2018-scidtb,braggaar2021creating} and Cohen's kappa \cite{cohen1960coefficient} for measuring inter-annotator agreement \cite{yang-li-2018-scidtb,nguyen-2018-bktreebank} are still widely used metrics for the annotation quality evaluation in the dependency syntax area.

\section{Annotation Task}
\label{sec:task}

\begin{figure*}[ht]
\begin{center}
\includegraphics[width=\hsize]{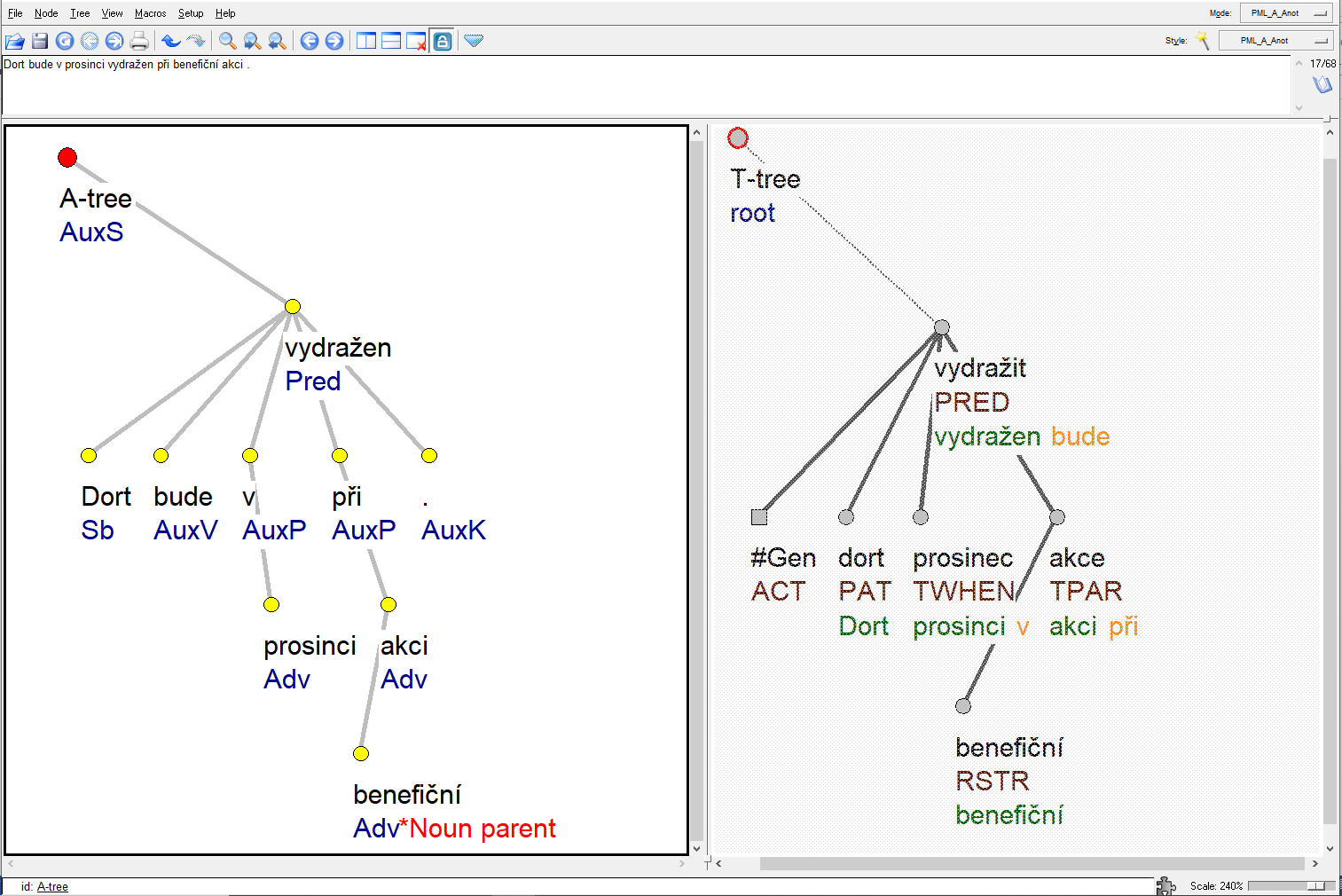} 
\caption{Annotation tool TrEd: Dependency syntax annotation of the sentence \textit{Dort bude v prosinci vydražen při benefiční akci} ‘The cake will be auctioned at a charity event in December.’ (on the left) with syntactic-semantic annotation of the same sentence available in a separate window (on the right), displayed in the stylesheet available to annotators during the annotation. In the dependency syntax annotation, the output of the automatic ``online'' checking procedure is also shown: there is an error in the incorrectly assigned afun {\tt Adv} to a node that depends on the noun. According to the annotation guidelines, any node that depends on a noun gets the afun {\tt Atr}.}
\label{fig:tred}
\end{center}
\end{figure*}


The experiment focuses on annotation at the dependency syntax layer within the multi-layered treebank for Czech -- PDT-C. Its version 1.0 features a fully manual morphological and syntactic-semantic annotation \cite{hajic-etal-2020-prague}. For version 2.0, the goal is to extend the fully manual mid-level dependency syntax annotation to all parts of PDT-C (this assumes an annotation of 2,000,000 tokens). For our experiments described in this paper, we have used a sample from the PCEDT part of PDT-C (Czech translation of the Wall Street Journal portion of Penn Treebank \citelanguageresource{penntb-LR}).

A dependency syntax annotation usually consists of determining the head for each node and assigning a syntactic function (attribute \textit{afun}) describing the relation between the dependent node and its head. In PDT-style dependency syntax annotation, every word (token) of a sentence (including punctuation marks) is represented by a node of the tree and at the same time, no added nodes are allowed. The original style of this layer annotation is best described in \newcite{analmanual}\footnote{\url{https://ufal.mff.cuni.cz/pdt-c/publications/PDT20-a-man-en.pdf}} based on principles first formulated in \newcite{biblio:HaBuildinga1998}.

The rules follow (wherever possible) traditional principles of the Czech grammar (especially as described in \newcite{vsmilauer1947novovceska}). In a prototypically structured sentence, a predicate ({\tt Pred}) is the head of the sentence and it depends on the technical root.
Afuns (25 types are distinguished) primarily describe dependency syntax relations such as subject ({\tt Sb}), object ({\tt Obj}), adverbial ({\tt Adv}), attribute ({\tt Atr}). Also all other nodes are assigned a head and a dependency relation (sometimes only in a purely technical way), e.g. auxiliary verbs ({\tt AuxV}), prepositions ({\tt AuxP}), subordinate conjunctions ({\tt AuxC}), coordination ({\tt Coord}) and apposition ({\tt Apos}) conjunctions, particles ({\tt AuxZ}) and punctuation ({\tt AuxG}, {\tt AuxX}, {\tt AuxK}).

Additional syntactic features such as ellipsis ({\tt \_E}) and parenthesis ({\tt \_P}) and members of coordination ({\tt \_Co}) or apposition ({\tt \_Ap}) structure are denoted by a specific affixes attached to appropriate afun. These three types of affixes can combine with each other, increasing the total number of possible afuns to 200.\footnote{Some combinations of afun and affix are in principle unacceptable, but the annotation tool (Sect.~\ref{sec:tools}) allows to create any of them, so when measuring the results of our experiment (Sect.~\ref{sec:results}), we take into account both the number of basic afuns (25) and the number of all possible afuns ($25 * 2^3 = 200$). Unacceptable combinations of afun and affix are then subject to automatic checks (Sect.~\secref{sec:cheks}).
Also, it should be noted that the rules for some afuns, ellipsis and affixes differ slightly from the original guidelines as referenced above.}  

An example of dependency syntax annotation displayed in the annotation tool (Sect.~\ref{sec:tools}) is shown in Fig.~\ref{fig:tred} (compared to syntactic-semantic annotation of the same sentence; read more in Sect.~\ref{sec:tr}).

\section{Annotation Tools Used}
\label{sec:tools}

For manual annotation, we use the TrEd annotation tool \cite{biblio:PaStRecentAdvances2008}. The basic editing capabilities of TrEd allow the user to easily modify the tree structure with drag-and-drop operations. The annotation process is greatly accelerated by a set of functions called macros (activated by keyboard shortcuts), customized to simplify the most common tasks done by the annotators (e.g., in our annotation task (Sect.~\ref{sec:task}), macros are available for assigning any of the afun values). TrEd provides many tools to customize the editor for various tasks. When annotating dependency syntax, we also utilize the possibility to write macros that incorporate consistency tests into the annotation process (Sect.~\ref{sec:cheks}). TrEd can also display more than one annotation (more than one tree) in one editing window. An annotator can thus check his/her annotation against another (Sect.~\ref{sec:tr}). The  annotation tool TrEd as we use it for the annotation task described here is shown in Fig.~\ref{fig:tred}.

\subsection{Pre-annotation}
\label{sec:pre-annotation}

To perform automatic pre-annotation, we selected a dependency parser achieving the best published performance on the Prague Dependency Treebank (PDT). To our best knowledge, that is the \textit{UDPipe 2} dependency parser combined with RobeCzech~\cite{straka-etal-2021-robeczech}, where RobeCzech is a Czech variant of RoBERTa.

We trained the parser on the PDT-style syntax including the afun affixes, using the train portion of PDT-C 1.0 \citelanguageresource{lrPDT-C}, reaching UAS 94.18\% and LAS 91.37\% on the PDT evaluation test set (the {\it etest} section). To ascertain the effect of out-of-domain data, we also evaluated the parser performance on 2,000 manually annotated sentences from PCEDT part of PDT-C, obtaining UAS 93.25\% and LAS\footnote{Please note that the LAS score referred to here is using a full set of the original afuns, including the original affixes, such as {\tt\_{}Co}; later in this paper, we define LAS differently and add one more metric (Sect.~\ref{sec:accuracy}).} 90.93\%, close to the PDT result despite the change of domain.

\subsection{Annotation Checking Rules}
\label{sec:cheks}

To get a higher quality and a greater consistency of annotated data, a set of automatic ``online'' checking procedures has been proposed and created in accordance with the annotation guidelines and incorporated into the annotation process to prevent the annotator from making accidental mistakes, e.g. because of poor attention. 
The ``online'' checks are based on the fact that many annotation rules imply that particular phenomena cannot (or have to) occur in the annotation output. They mainly work with the combination of attribute values and the structure of a tree. In our experiments, 9 sets of automatic checking rules were available.

Some of the checking rules utilize the multi-layered structure of the PDT-C and use information from the lower morphological layer (e.g. morphological categories, lemmas). For example a rule states that an attribute (afun {\tt Atr}) never depends on a verb and a node that depends on the noun is an attribute (afun {\tt Atr}), or a rule states that a nominal part of a predicate (afun {\tt Pnom}) always depends on a verb \textit{být} ‘to be’ and its variants.

Many checks combine several rules, for example a check examines whether the afun {\tt Pred} for predicate is a head of the tree, except cases when it is denoted as a parenthesis, or a member of coordination or apposition by affixes {\tt \_P}, {\tt \_Co}, or {\tt \_Ap}. Further checking rules are defined by enumeration, for example a check states that the head of a tree has only a limited set of possible afuns: {\tt Coord} for coordination head, {\tt Apos} for apposition head, {\tt Pred} for predicate, {\tt Denom} for noun phrase, {\tt Partl} for interjection. Also some combinations of afun and affix are excluded, for example the punctuation marks {\tt AuxX} and {\tt AuxG} are never complemented with any affixes.

It is recommended to the annotators to run the checks after annotation of every single tree and consequently check and fix possible errors. 
In Fig.~\ref{fig:tred}, we can see an example of output of the automatic checks. The automatic ``online'' checking procedure revealed an error in the incorrectly assigned afun {\tt Adv} to a node that depends on the noun. As mentioned above, according to the annotation guidelines, any node that depends on a noun gets the afun {\tt Atr}.

\subsection{Availability of Other Annotation}
\label{sec:tr}

The dependency syntax annotation can take advantage of the availability of other manual annotation on the same data. Especially the higher, syntactic-semantic layer manual annotation with its tree structure can serve as a guidance because both layers are derived from the same syntactic theory developed in the framework of Functional Generative Description \cite{Sgall1986}. 

The main difference between the two syntactic layers lies in the fact that at the dependency syntax layer, every word of a sentence is represented by a node of its own and at the same time, no added nodes are allowed (cf. Sect.~\secref{sec:task}), while at the syntactic-semantic layer, only the content words are represented by a separate node and the function words such as prepositions, conjunctions, auxiliary or modal verbs do not have their own node: their meaning is captured by attributes of the nodes for the respective content words. Moreover, the cases of surface deletion are resolved by inserting new nodes into the tree structure.\footnote{Detailed annotation rules can be found in the annotation guidelines \cite{trmanualanot,biblio:MiAnnotationtectogrammatical2014}.}

As an illustration, note the differences between the two layers in Fig.~\ref{fig:tred} depicting annotation of the sentence \textit{Dort bude v prosinci vydražen při benefiční akci.} ‘The cake will be auctioned at a charity event in December’.\footnote{The TrEd concept of stylesheets allows to visually differentiate nodes, edges by color, shape, etc. and also visualize cross-layer relations. The trees in Fig.~\ref{fig:tred} are shown in the stylesheet available to annotators during the annotation. Under a lemma in the syntactic-semantic tree (on the right side), the function words belonging to the given node are displayed in orange.} Fig.~\ref{fig:tred} shows that at the dependency syntax layer, the auxiliary verb \textit{bude} ‘will’ has its own node with label {\tt AuxV}, while at the syntactic-semantic layer, its function is captured  within the node for the predicate. It is similar for prepositions (\textit{v} ‘in’, \textit{při} ‘at’) that have their nodes (with label {\tt AuxP}) at the dependency syntax layer, but not at the syntactic-semantic one. 
Otherwise, there is an added node for the so called general actor (with lemma \textit{\#Gen}) at the syntactic-semantic layer, which is not expressed in the surface shape of the sentence. The semantic labels such as {\tt ACT} (Actor), {\tt PAT} (Patient), {\tt TWHEN} (When) and {\tt RSTR}  (Restriction) usually correspond to afuns {\tt Sb}, {\tt Obj}, {\tt Adv} and {\tt Atr} respectively at the dependency syntax layer. However, there are several exceptions such as in case of passive constructions, etc.

Annotators see both trees side by side during the annotation, so they can check their annotation against the other one.\footnote{The annotators do not have to have a full knowledge of the other annotation structure and guidelines; it is sufficient they know the principles in order to be able to ``read'' the annotation as shown in TrEd.} Although the principles of the annotation on both layers are different and the resulting tree structure may also be very dissimilar, we assume that the availability of (largely) semantic annotation during the dependency syntax annotation speeds up the understanding of a sentence structure (especially in the case of long sentences), helps in ambiguous and complicated cases, both in solving the syntactic structure and in determining afuns; and finally increases the consistency of the annotated data.

\section{Experiment} 
\label{sec:experiment}

In this section, we describe the experiment, its design (Sect.~\secref{sec:design}), results (Sect.~\secref{sec:results}) and evaluation (Sect.~\secref{sec:evaluation}). Input data, individual experimental annotations, and a complete and detailed overview of the measured results are freely available (as a supplemental material to this paper) via the LINDAT/CLARIAH-CZ repository \citelanguageresource{lrexperiment}.\footnote{\url{http://hdl.handle.net/11234/1-4647}}

\subsection{Design}
\label{sec:design}

In line with the research questions that we want to address and the annotators that we have available, we choose the following experiment design. 

In the experiment, we examine four annotation tasks: annotation with no support (abbreviated \textit{no\_supp}), annotation with checking rules available (abbreviated \textit{rules}; cf. Sect.~\ref{sec:cheks}), annotation with syntactic-semantic annotation on the same data available (abbreviated \textit{annot}; cf. Sect.~\ref{sec:tr}), and annotation with checking rules and syntactic-semantic annotation on the same data available (abbreviated \textit{rul\_annot}). For each task, the data was prepared in two modes: annotation on pre-parsed data (abbreviated \textit{pre-parsed}; cf. Sect.~\ref{sec:pre-annotation}) and annotation ``from scratch'' (abbreviated \textit{from-scratch}). In the experiment, we thus examined eight different annotation set-ups (four tasks in two modes).

\begin{table}[ht]
\scriptsize
\centering
\begin{tabular}{l|ll|ll|ll|ll}
Task & \multicolumn{2}{l|}{no\_supp}  & \multicolumn{2}{l|}{rules}  & \multicolumn{2}{l|}{annot} & \multicolumn{2}{l}{rul\_annot}\\\hline
Data & D1   & D2 & D3   & D4 & D5 & D6 & D7 & D8 \\\hline
Pre- & a1   & a3 & a1   & a3 & a1 & a3 & a1 & a3 \\
parsed & a2   & a4 & a2   & a4 & a2 & a4 & a2 & a4 \\\hline
From- & a3   & a1 & a3   & a1 & a3 & a1 & a3 & a1 \\
scratch & a4   & a2 & a4   & a2 & a4 & a2 & a4 & a2 \\
\end{tabular}
\caption{Data distribution}
\label{tab:data-distribution}
\end{table}

We prepared eight datasets (\textit{D1} -\textit{ D8}) of 1,250 words, two sets for each task, i.e., a total of 2,500 words per task. We prepared each dataset in both modes (for pre-parsed and from-scratch annotation).  The datasets were selected to be as similar as possible in terms of the number of sentences (each dataset contains approximately 60 sentences), number of nodes per sentence, depth of tree, etc. Each dataset was a coherent, continuous text. With this selection, we eliminated the data effect on the resulting values.

Each task was annotated by four annotators (abbreviated \textit{a1}, \textit{a2}, \textit{a3}, \textit{a4}) in stable (unchanged) pairs. The annotation pairs were composed so that they were experientially balanced. Each pair thus experiences all eight annotation set-ups and no annotator annotated any data twice (cf. Tab.~\ref{tab:data-distribution} of data distribution). With this setting, we eliminated the learning effect and reduced the impact of the experience and ``style'' of the annotators on the resulting (averaged) values.

For each dataset, a gold standard data annotation was also created. As follows from the data distribution (cf. Tab.~\ref{tab:data-distribution}), each dataset was annotated four times (twice in from-scratch mode and twice in pre-parsed mode). The annotations were then diff-ed and disagreements were resolved by the guideline setters.

\subsection{Results}
\label{sec:results}

\subsubsection{Measuring accuracy}
\label{sec:accuracy}

We measured the accuracy of the annotation against the gold data. We measured the unlabeled and labeled attachment score for each dataset in each task for each mode, separately for each annotator. For each annotation set-up (for each task in the given mode) we obtained four values (cf. a1 to a4 in each cell of Tab.~\ref{tab:data-distribution}), from which we then calculated the average value.

We distinguish two labeled attachment scores. The first score, abbreviated LAS, measures accuracy on 25 basic labels (afuns), ignoring the three types of affixes. Afuns including affixes (increasing the number of labels to 200, cf. Sect.~\secref{sec:task}) are taken into account in the LAS score variant abbreviated as FULL.

All three scores are calculated in the same way: the number of agreements divided by the total number of edges. 
The agreement means the same head for a node for UAS, the same head and the same afun for LAS, and the same head and the same afun including affixes for FULL. The results are presented in Tab.~\ref{tab:accuracy}.


\begin{table}[t]
\small
\begin{center}
\setlength{\tabcolsep}{5pt}
\begin{tabular}{l|l|l}
Task/Mode & \multicolumn{1}{c|}{Pre-parsed} & \multicolumn{1}{c}{From-scratch} \\\hline
\multicolumn{3}{c}{UAS}\\\hline
no\_supp & 96.5 ±0.46 & 96.5 ±0.48  \\
rules & 96.8 ±0.42 p=30.9\% & 95.9 ±0.50 p=79.2\%  \\
annot & 97.5 ±0.48 p=~~7.1\% & 97.0 ±0.48 p=24.5\% \\
rul\_annot &  97.6 ±0.37 p=~~3.5\% & 97.8 ±0.32 p=~~1.5\% \\\hline
\multicolumn{3}{c}{LAS}\\\hline
no\_supp & 95.0 ±0.55 & 94.5 ±0.55  \\
rules & 95.3 ±0.49 p=30.6\% & 93.9  ±0.59 p=74.9\% \\
annot & 96.5 ±0.58 p=~~3.3\% & 95.9 ±0.55 p=~~4.7\% \\
rul\_annot & 96.6 ±0.45 p=~~1.3\% &  96.8 ±0.38 p\rlap{\raisebox{1.2pt}{$\scriptstyle<$}}\phantom{=}~~0.1\% \\\hline
\multicolumn{3}{c}{FULL}\\\hline
no\_supp & 94.5 ±0.56 & 94.3 ±0.56  \\
rules & 95.1 ±0.50 p=23.6\% & 93.5 ±0.60 p=80.1\% \\
annot & 96.1 ±0.60 p=~~3.5\% & 95.2 ±0.63 p=14.4\% \\
rul\_annot & 96.4 ±0.47 p=~~0.8\% & 96.6 ±0.40 p\rlap{\raisebox{1.2pt}{$\scriptstyle<$}}\phantom{=}~~0.1\% \\
\hline
\end{tabular}
\caption{\textbf{Accuracy} of annotation in various set-ups; the standard deviations are estimated using bootstrap resampling with 1M samples, and $p$ stands for a p-value that the given configuration is statistically significantly better than the corresponding \textit{no\_supp} configuration, estimated by a Monte Carlo permutation test with 1M samples \protect\cite{fay-mc-permutation-test}.}
\label{tab:accuracy}
\end{center}
\end{table} 

\subsubsection{Inter-annotator agreement}
\label{sec:iaa}

Inter-annotator agreement among annotators is a standard way to determine consistency in the process of annotation. We use a widely used measure, Cohen's kappa, $\kappa$ \cite{cohen1960coefficient} for measuring agreement between annotators. Although there has been criticism of the inadequacy of this method for structure annotation \cite{skjaerholt-2014-chance}, it provides a standardized way of estimating the agreement rate between annotators. 

Similarly to Sect.~\ref{sec:accuracy}, we measured three different kappas: unlabeled, labeled, and full. But the details were quite different: for the traditional formula
$$
\kappa = 1 - {(1 - p_0) \over (1 - p_e)}
$$
the values of $p_0$ (actual agreement) and $p_e$ (probability of a random agreement) were calculated as follows:

\begin{itemize}

\item For unlabeled kappa, $p_0 = a_P / n$ where $a_P$ is the number of nodes with the same head in both annotations, while $n$ is the total number of nodes (as nodes are not added or removed, the number is always the same for both the annotators); $p_e = \bar{s}^{-1}$ where $\bar{s}$ is the average sentence size.

\item For labeled kappa, $p_0 = a_L / a_P$, i.e. we only consider the edges that are common to both the annotations and we check their labels ($a_L$ is the number of agreements on afuns); $p_e = 1 / 25$ where 25 is the number of all possible labels (afuns).

\item The full labeled kappa is similar: $p_0 = a_F / a_P$ and $p_e = 1 / 200$ (where $a_F$ is the number of agreements on labels including affixes). Clearly, the probability of a random agreement is very low for such a high number of labels.

\end{itemize}

We measured the inter-annotator agreement for each pair of annotators. For each task in the given mode, we obtained two values (cf. pairs a1-a2 and a3-a4 in each cell of Tab.~\ref{tab:data-distribution}), from which we then calculated the average values. The results are presented in Tab.~\ref{tab:iaa}.

\begin{table}[h]
\small
\begin{center}
\begin{tabular}{l|c|c}
Task/Mode &  Pre-parsed & From-scratch \\\hline
\multicolumn{3}{c}{Unlabeled $\kappa$}\\\hline
no\_supp & 0.96   &  0.95 \\
rules & 0.97 & 0.94  \\
annot  & 0.97 & 0.96 \\
rul\_annot  & 0.97  & 0.97  \\\hline
\multicolumn{3}{c}{Labeled $\kappa$}\\\hline
no\_supp & 0.99 & 0.96  \\
rules & 0.98 & 0.97 \\
annot & 0.99 & 0.98 \\
rul\_annot  & 0.99 & 0.99 \\\hline
\multicolumn{3}{c}{Full-labeled $\kappa$}\\\hline
no\_supp & 0.98 & 0.96  \\
rules & 0.98  & 0.97 \\
annot & 0.98  & 0.97 \\
rul\_annot & 0.98  & 0.99  \\
\hline
\end{tabular}
\caption{\textbf{Agreement rate} in various set-ups}
\label{tab:iaa}
\end{center}
\end{table} 

\subsubsection{Annotation time}
\label{sec:time}
In Tab.~\ref{tab:time}, we report the total time taken in annotation for each annotation set-up of each annotator. 

\begin{table}[t]
\small
\begin{center}
\begin{tabular}{l|l|c|c}
\multicolumn{2}{l|}{Task/Mode} & Pre-parsed & From-scratch \\\hline
no\_supp & a1 & 66 & 150  \\
  & a2 & 125 & 231  \\
  & a3 & 80 &  140 \\
  & a4 & 200 &   280\\\hline
rules &  a1 & 90 & 156  \\
  & a2 & 216 & 252  \\
  & a3 & 60 & 150  \\
  & a4 & 150 &  180 \\\hline
annot & a1 & 111  & 130 \\
  & a2 & 212 & 205  \\
  & a3 & 70 & 180  \\
  & a4 & 210 & 230  \\\hline
rul\_annot & a1  & 80  & 150 \\
  & a2 & 200 & 245  \\
  & a3 & 50 & 160  \\
  & a4 & 180 & 155  \\
\hline
\end{tabular}
\caption{\textbf{Time} in various set-ups (in mins)}
\label{tab:time}
\end{center}
\end{table} 

\subsection{Evaluation}
\label{sec:evaluation}

In this section, we judge the results of the experiment (presented above in Sect.~\ref{sec:results}) with respect to the research questions stated in the Sect.~\ref{sec:intro}.

\subsubsection{Does annotation quality remain acceptably high if  pre-annotation is used?}

To answer the main research question, we compare the values obtained for the pre-parsed and for from-scratch data for accuracy (cf. Tab.~\ref{tab:accuracy}), inter-annotator agreement (cf. Tab.~\ref{tab:iaa}), and annotation time  (cf. Tab.~\ref{tab:time}). 

\begin{figure}[ht]
\begin{center}
\includegraphics[width=\hsize]{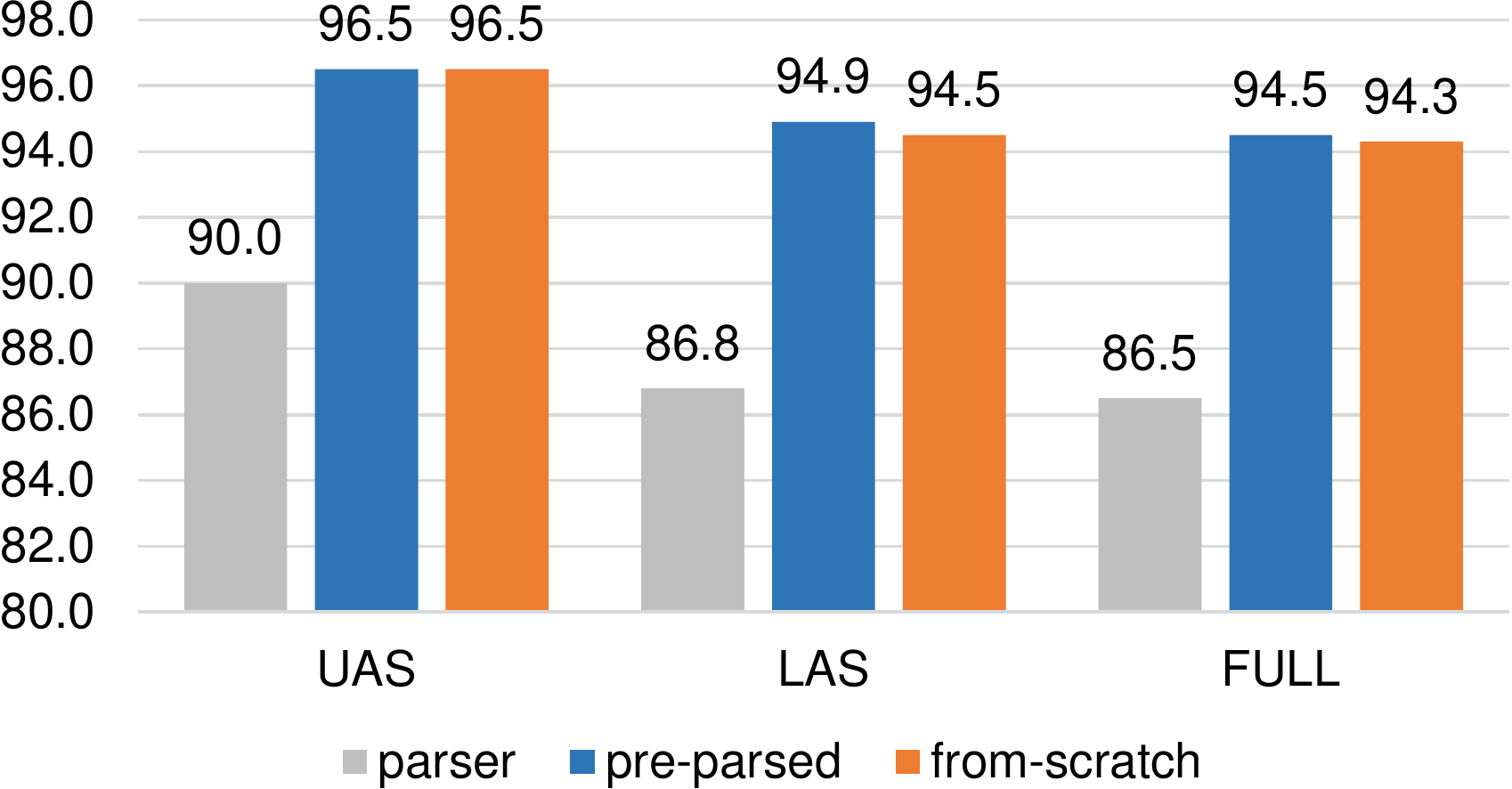} 
\caption{\textbf{Accuracy} of the annotation of no-support task in comparison with accuracy of the parser}
\label{fig:accuracy-overall}
\end{center}
\end{figure}

From Tab.~\ref{tab:accuracy}, it is clear that the annotation accuracy remained, more or less, the same for both modes of annotation, with a fluctuation (as measured by the standard deviation) of around 0.5 percentage points. 
Cf. also Fig.~\ref{fig:accuracy-overall} which depicts overall accuracy only on the annotation of no-support task.

To find out how much manual correction of automatic pre-annotation increases the quality, we also measured the accuracy of the parser used (i.e., the accuracy of the input data for tasks in "pre-parsed" mode) against the gold data. An overview of the obtained values for parser, annotation on pre-parsed data (no-support task) and annotation on from-scratch data (no-support task) is shown in Fig.~\ref{fig:accuracy-overall}. 

Lower parsing accuracy (compared to the values stated in Sect.~\ref{sec:pre-annotation}) are due to the fact that we have changed the domain and also some annotation rules (cancellation of some afuns, addition of new ones, etc., even if only a small minority, occurrence-wise) compared to the previous dataset on which the parser has been trained. The graph shows that manual annotation (whether on pre-parsed data or not) produces higher quality data than the output from the parser.

\begin{figure}[t]
\begin{center}
\includegraphics[width=\hsize]{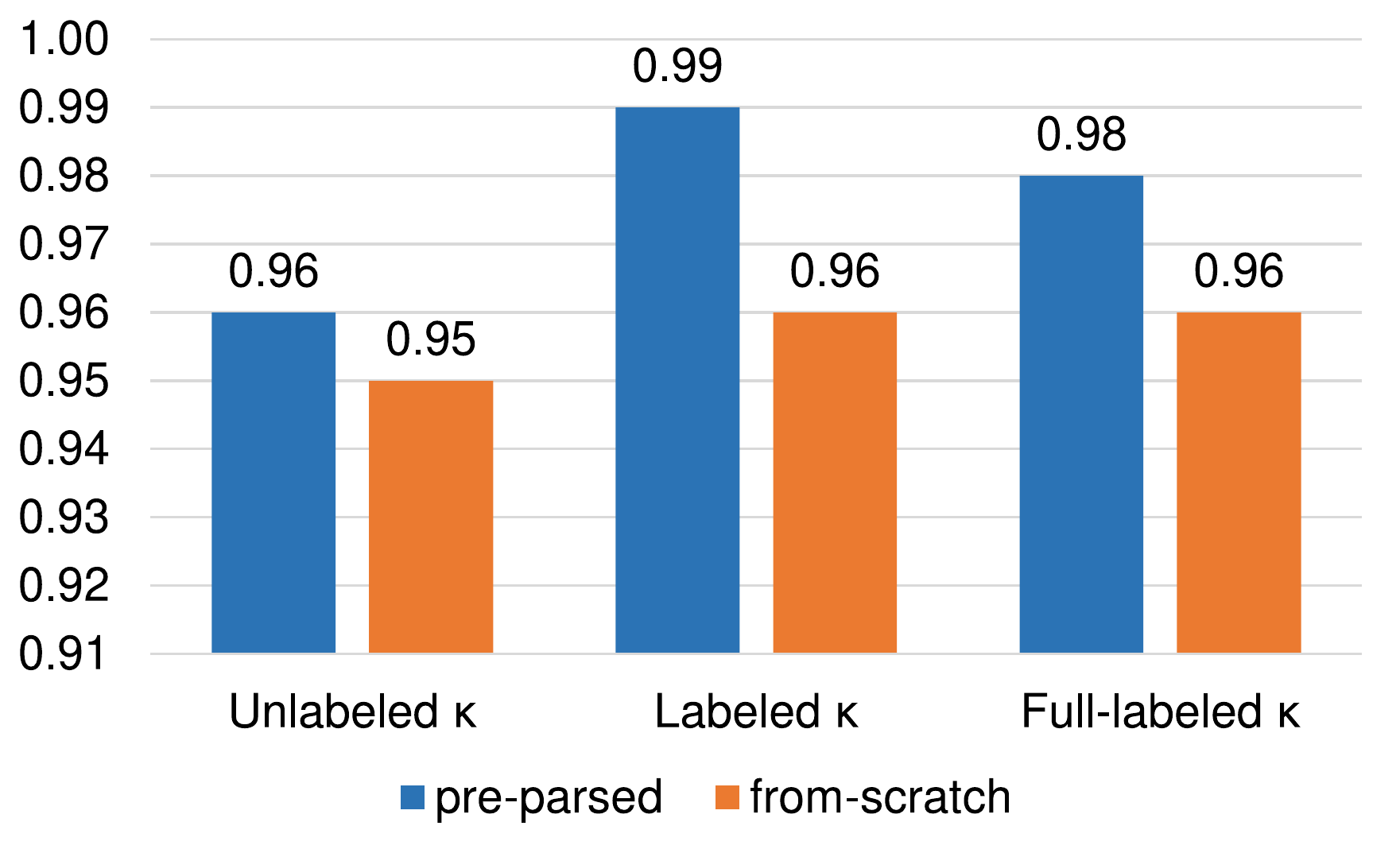} 
\caption{\textbf{Agreement rate} of no-support annotation}
\label{fig:kappa-overall}
\end{center}
\end{figure}

Fig.~\ref{fig:kappa-overall} depicts total agreement rate for annotation on the no-support task. From the Fig.~\ref{fig:kappa-overall}, it is obvious that the pre-annotation contributes to the consistency of the annotation. This can be attributed to the fact that judgments of annotators are influenced by the output of automatic pre-annotation.
However, as stated above, pre-annotation does not affect the quality of the result.


We can also observe that the variance in the annotation speed between individual annotators is large, from 50 to 200 minutes per task of pre-parsed data and from 130 to 280 minutes per task of from-scratch data  (cf. Tab.~\ref{tab:time}).
On average, the annotation on from-scratch data takes almost 1.7 times longer than annotation on pre-parsed data. 

It follows from these results that the answer to the first research question is in the affirmative:
the quality of the annotation is not hurt by the pre-parsing mode while the speed is up and inter-annotator agreement increases.

\subsubsection{What effect do the used tools have on annotation quality and efficiency?}
\label{sec:answer2}

To answer the question how the quality and speed of annotation changes when using other support tools, we compare in detail the figures obtained for each task in Tables~\ref{tab:accuracy}, \ref{tab:iaa}, and \ref{tab:time}.



As mentioned in the previous section, the annotation accuracy remained, more or less, the same for both modes. As can be seen from the figures in Tab.~\ref{tab:accuracy},
we can observe how much the addition of a support tool \textit{(rules}, \textit{annot}, \textit{rul\_annot}) has increased the quality of annotation. While the differences seem small, some of them are significant when compared to the \textit{no\_supp} task (see the $p$-values in Tab.~\ref{tab:accuracy}). Detailed insight through the significance tests shows that especially the combination of the ``online'' checking rules and the possibility to easily visually check the other type of annotation (the syntactic-semantic one) and make changes under its guidance (line \textit{rul\_{}annot} in Tab.~\ref{tab:accuracy}) did improve the quality of the annotation significantly for all of UAS, LAS, and FULL metrics, in some cases very convincingly (as demonstrated by a very small $p$-value). Even the use of the other annotated data alone (line \textit{annot} in Tab.~\ref{tab:accuracy}) improved some metrics in at least one mode significantly, even if less convincingly: LAS for both the pre-parsed and from-scratch mode, and FULL for the pre-parsed mode; UAS for the pre-parsed mode lies on the boundary of acceptable significance.
 
The apparently small influence when using the checking rules as the only support (line \textit{rules} in Tab.~\ref{tab:accuracy}) can be attributed to the fact that there were relatively only a few sets of checks available at the time of the experiment. The more interesting it is then to observe that the combination of the ``online'' checking rules and the availability of the other annotation (\textit{rul\_{}annot}) has increased the quality so convincingly.



Regarding the consistency of annotation, the figures in Tab.~\ref{tab:iaa}
show the total agreement rate for each task on pre-parsed and from scratch data. We can observe that the use of annotation at the syntactic-semantic layer as a support lead to higher consistency of the annotation, especially on from-scratch input data. If pre-annotation is used, the effect of other tools (including other annotation support) on the consistency of annotation is unclear. 

\begin{figure}[ht]
\begin{center}
\includegraphics[width=.9\hsize]{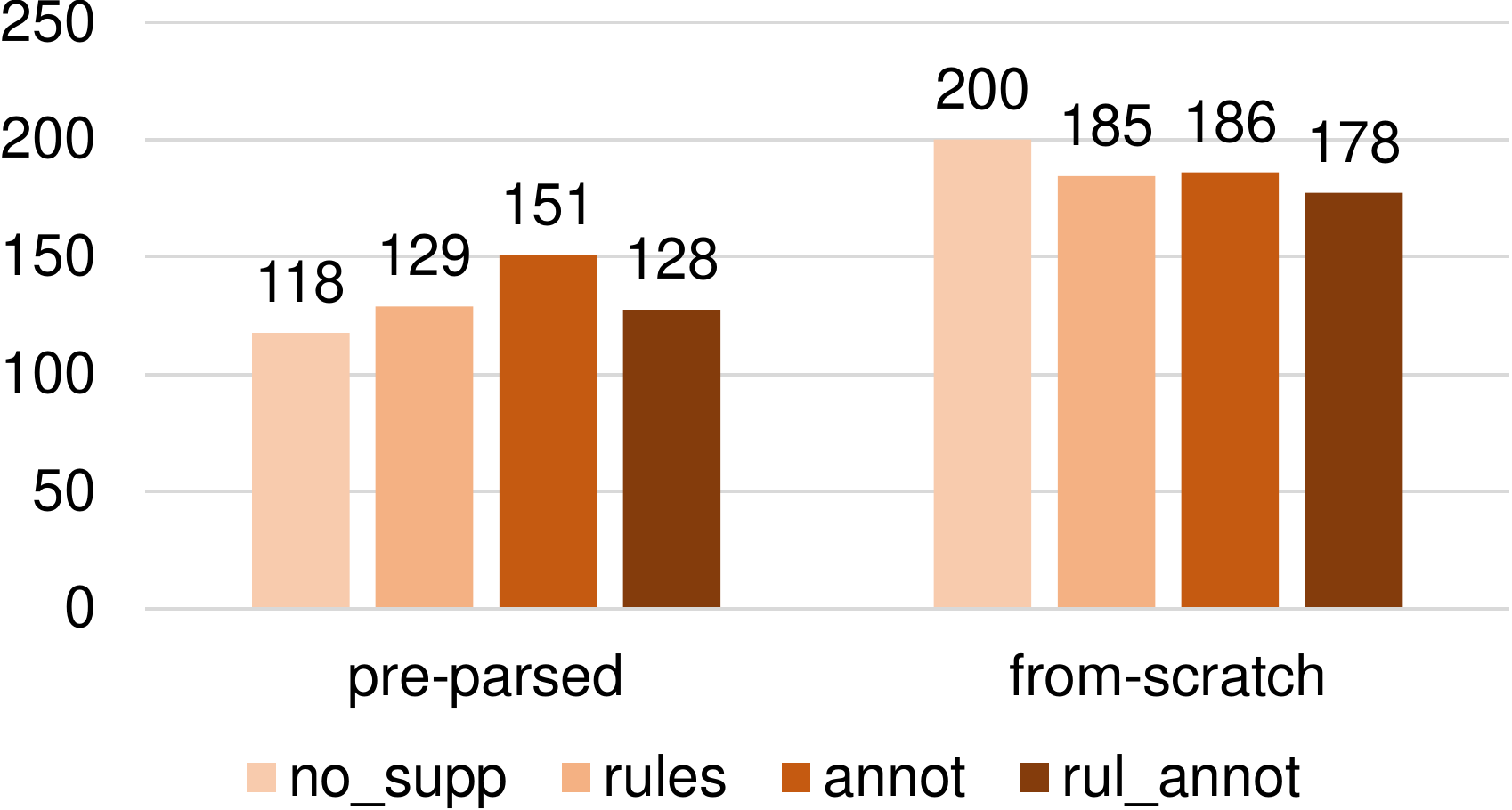} 
\caption{\textbf{Time} of annotation}
\label{fig:time}
\end{center}
\end{figure}

It is hard to draw any conclusions regarding the correlation of speed of the annotation and the use of the support tools, due to the high variation of the speed of annotation across the annotators. 
But perhaps this is in fact good news: the annotation time does not increase (too much) due to the use of checking rules, neither when annotators have the other annotation available to check against (cf. Tab.~\ref{tab:time} and  detailed graph in Fig.~\ref{fig:time}).

\subsubsection{Which set-up is most useful for manual dependency syntax annotation?}

The experiment confirms the usefulness of automatic pre-annotation for efficiency of annotation and quality of the result. It significantly increases the speed of annotation and does not reduce the quality of data.

The influence of the support tools is not as positive as expected, especially when using the checking rules alone. This contradicts our previous experience (e.g. \newcite{biblio:MiStWaysEvaluation2010}). Perhaps it will help to develop more checking rules throughout the annotation project, once it starts in full. Regarding the use of annotation at the syntactic-semantic layer, the agreement rate figures measured on the from-scratch data show that the other annotation support has the potential to increase the consistency of the annotation, as well as the quality if used in conjunction with the checking rules. 
Therefore, and because these tools do not increase annotation time too much (cf. Fig.~\ref{fig:time}), we will use them in the annotation project proper (together with pre-parsing).

\begin{figure}[ht]
\begin{center}
\includegraphics[width=\hsize]{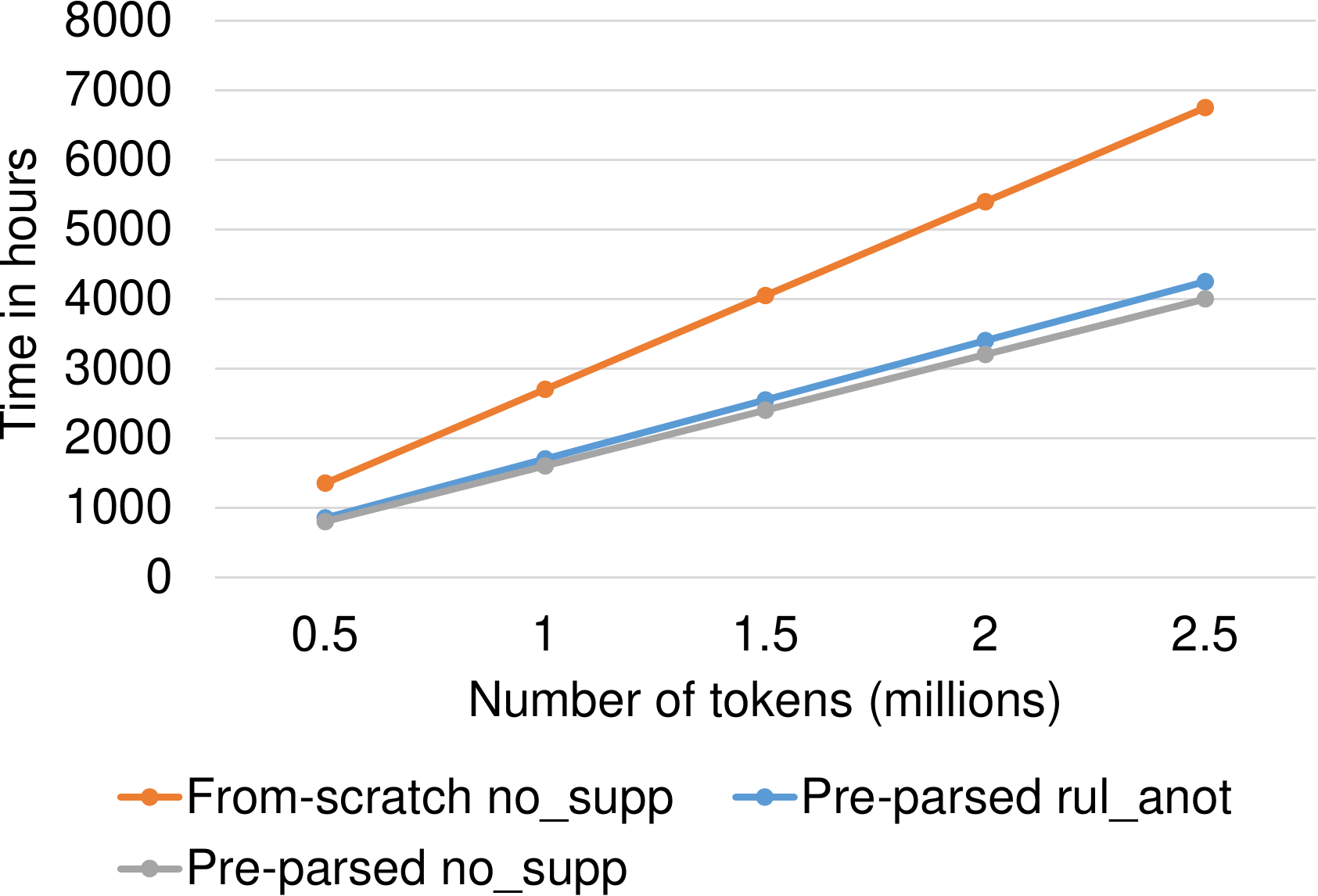} 
\caption{\textbf{Extrapolated hours} of annotation needed}
\label{fig:timesaved}
\end{center}
\end{figure}

A graph in Fig.~\ref{fig:timesaved} depicts the extrapolated time for annotation using just manual work (from scratch with no annotation support) vs.~the time needed when using pre-parsed data (with and without the additional annotation support) as presented in this paper. We leave the calculation of financial savings to the readers, based on their usual hourly rates. Please note that any hourly savings of the primary annotation induce additional savings of technical support, supervision etc., which is however harder to quantify. 

In our case, for the PDT-C corpora annotation with a total of about 2 million tokens, the primary saving will be approx. 2,000 hours (5,400--3,400) per single annotation pass (\~{}1.3 person-years, or at least about \${}21,000) if we compare the annotation with no support and the annotation with all available support; with the usual double annotation, it is then also double the savings in both time and cost.

\section{Conclusions and Future Work}
\label{sec:conclusions}

In the paper, we presented pre-annotation bias experiment for a mid-level annotation complexity task--dependency syntax annotation within the Prague Dependency Treebank - Consolidated 2.0 project which is now ongoing. We judged efficiency and quality of annotation performed under various experimental set-ups: annotation with no support (``from-scratch'') and annotation supported by the use of automatic pre-annotation (using a high-accuracy parser), availability of other annotation on the same data, and ``online'' checking rules (implemented in such a way that the annotator can check the manual annotation immediately). 

The experiment confirmed that pre-annotation using UDPipe 2 is an efficient tool for manual syntactic annotation, which increases the speed and consistency of the resulting annotation without reducing its quality. Thus we can conclude it is highly recommended to use automatic pre-annotation in syntactic annotation projects, even if the other tools are not available.

The influence of the other support tools (online annotation checking rules and the availability of a related, even if largely not directly comparable syntactic-semantic annotation on the same data) on efficiency and quality of annotation was not as positive as expected. However, when used together they significantly increase the quality of the annotation, in both the pre-parsed and from-scratch modes. As these tools do not increase annotation time (much), we will use them in the upcoming large-scale PDT-C dependency syntax annotation process. 

We plan to repeat the experiment at half-time of the project to re-evaluate the influence of the chosen annotation set-up on the quality and efficiency of the annotation.

\section{Acknowledgements}
We thank the annotators for their commitment in the implementation of the experiment.

The research and language resource work reported in the paper has been supported by the LINDAT/CLARIAH-CZ project funded by Ministry of Education, Youth and Sports of the Czech Republic (project LM2018101) and by the EXPRO project LUSyD, funded by the Grant Agency of the Czech Republic as project No. GX20-16819X.

\section{Bibliographical References}\label{reference}

\bibliographystyle{lrec2022-bib}
\bibliography{lrec2022-annotation}

\section{Language Resource References}
\label{lr:ref}
\bibliographystylelanguageresource{lrec2022-bib}
\bibliographylanguageresource{languageresource}

\end{document}